\documentclass{article}

     \usepackage[final, nonatbib]{mlwg_2019}

\usepackage{hyperref,times,helvet,courier,subfig,dblfloatfix,setspace,ifthen,amsmath, amsthm, amssymb, graphicx,amsfonts,multirow,algorithm,algorithmic,float}
\usepackage[utf8]{inputenc} 
\usepackage[T1]{fontenc}    
\usepackage{hyperref}       
\usepackage{url}            
\usepackage{booktabs}       
\usepackage{amsfonts}       
\usepackage{nicefrac}       
\usepackage{microtype}      
\newtheorem{theorem}{Theorem}
\newtheorem{lemma}[theorem]{Lemma}
\newtheorem{definition}{Definition}

\title{Semisupervised Representation Learning based on Probabilistic Labeling with Performance Guarantee}

\author{%
  Ershad Banijamali\\
  School of Computer Science\\
  University of Waterloo\\
  Vector Institute\\
  \texttt{sbanijam@uwaterloo.ca} \\
   \And
   Ali Ghodsi \\
   Department of Statistics and Actuarial Science \\
   University of Waterloo \\
   \texttt{aghodsib@uwaterloo.ca} \\
}

\begin{document}

\maketitle
\maketitle
\begin{abstract}
In this paper, we present a new algorithm (SSRL-PL) for semi-supervised representation learning. In this algorithm, we first find a vector representation for the labels of the data points based on their local positions in the space. Then, we map the data to lower-dimensional space using a linear transformation such that the dependency between the transformed data and the assigned labels is maximized. In fact, we try to find a mapping that is as discriminative as possible. The approach will use Hilber-Schmidt Independence Criterion (HSIC) as the dependence measure. We also present a kernelized version of the algorithm, which allows non-linear transformations and provides more flexibility in finding the appropriate mapping. Use of unlabeled data for learning new representation is not always beneficial and there is no algorithm that can deterministically guarantee the improvement of the performance by exploiting unlabeled data. Therefore, we also propose a bound on the performance of the algorithm, which can be used to determine the effectiveness of using the unlabeled data in the algorithm. We demonstrate the performance of the algorithm using both toy examples and real-world datasets.
\end{abstract}
\section{Introduction}
As the amount of data grows rapidly, the process of extracting meaningful information becomes more and more challenging. In the real world, the amount of labeled data compared to unlabeled data is almost negligible. On the other hand, determining data categories, or acquiring labels, is expensive for many reasons, e.g. it is extremely time-consuming for large datasets and usually needs human supervision.
Semi-supervised learning is the area of utilizing unlabeled data combined with, usually very smaller set of, labeled data to gain better data representation or classification accuracy.  

\textbf{Prior Art: }In recent years, semi-supervised learning has attracted attention from many researchers and several algorithms have been designed for semi-supervised learning that can relate to the present work. Graph-based algorithms, which usually define a loss function for labeled data and use unlabeled as a regularizer, are important classes of semi-supervised learning methods. Example of this class are \cite{blum2001,zhu2003} that try to convey the label information over the edges of the graph. Label propagation has been tried in many other articles including \cite{zang2012} which, inspired by the idea of locally linear embedding (LLE) \cite{roweis2000},  assumes the labels of data points can be linearly constructed by the labels of their adjacent samples in an sparse neighborhood and \cite{zhang2015}, which tries to propagate the labels over pairs of data points. Transductive support vector machines (TSVM) is another class of algorithms, used by \cite{chapelle2005}, in which the goal is to maximize the margin for both unlabeled and labeled points. 

Unlike off-the-shelf feature extraction, \cite{nouredanesh2016gabor, nouredanesh2016radon}, there exists no guarantee  in semi-supervised learning that the use of unlabeled data will help us to achieve a better representation of the data \cite{cozman2003}. To reduce the likelihood of having destructive unlabeled samples, there is a set of assumptions about the structure of the distribution of data, including smoothness assumption, clustering assumption, and manifold assumption.

\textbf{Contribution:} Most of the semi-supervised algorithms include two objective functions for labeled and unlabeled data points, which are optimized jointly. In this paper, we also start with deriving two separate objective functions. For the labeled points, we look for a mapping which maximizes the dependency of the transformed points and their labels, and for the unlabeled points we look for a mapping that keeps them near their labeled neighbors. However, by some manipulations, we then combine these two functions and solve the problem by optimizing a single objective function. Further investigations show that the objective function can also be obtained by a specific assignment of labels to the points. We call this probabilistic labeling. This labeling not only provides the objective function of our problem much faster and easier, but also enables us to obtain a bound on the performance of the algorithm based on probability of classification error in the original space. This bound  shows the maximum deviation of the objective function  value from its optimal value, when we know the true label of all data points in our dataset.
We will also present the kernelized version of the algorithm, which is helpful when the linear transformation does not provide a good representation of data in the target space. 
\section{Background: Hilber-Schmidt Independence Criterion (HSIC)}
The Hilbert-Schmidt Independence Criterion (HSIC) is a very useful tool in statistics to measure the dependence between two random variables \cite{gretton2005}. We use HSIC in our proposed method. Following is a short description about this measure.
\begin{definition}
Suppose $\mathcal{X}$ and $\mathcal{Y}$ are two domain sets. Let $\phi$ and $\psi$ be two mappings that map $\mathcal{X}$ and $\mathcal{Y}$ to their corresponding Reproducing Kernel Hilbert Space (RKHS) $\mathcal{F}$ and $\mathcal{G}$. The Borel probability measure over  $\mathcal{X}\times \mathcal{Y}$ is denoted by $p_{xy}$. Then HSIC is defined as the following:
\begin{equation}
\text{HSIC}(p_{xy},\mathcal{F},\mathcal{G}) = \parallel \textbf{E}_{x,y}[\phi(x) \otimes \psi(y)] - \mu_x \otimes \mu_y \parallel_{HS}^2
\end{equation}  
where $\mu_x$ and $\mu_y$ are mean of $\phi(x)$ and $\psi(y)$, respectively, and  $\otimes$ is the tensor product. $\parallel . \parallel_{HS}$ is also the Hilbert-Schmidt norm.
\end{definition}

The following theorem by \cite{gretton2005} shows the relation between HSIC and independence of $x$ and $y$, when $(x,y)$ is drawn from $p_{xy}$.

\begin{theorem}
Suppose $k$ and $l$ are reproducing kernels of RKHS's $\mathcal{F}$ and $\mathcal{G}$ on the compact domains $\mathcal{X}$ and $\mathcal{Y}$. Assume, without loss of generality, $\parallel f \parallel_{\infty} \leq 1$ and $\parallel g \parallel_{\infty} \leq 1$ for all $f\in \mathcal{F}$ and $g\in \mathcal{G}$. Then, $\text{HSIC}(p_{xy},\mathcal{F},\mathcal{G})$ is zero, if and only if, $x$ and $y$ are independent.
\end{theorem}

\subsection{Empirical HSIC}
The empirical HSIC was also defined in \cite{gretton2005} to show that HSIC is, in fact, a practical criterion.
\begin{definition}
Let $Z = \{(x_1,y_1),(x_2,y_2),...,(x_m,y_m) \subseteq \mathcal{X}\times \mathcal{Y}$ be a series of $m$ independent observation drawn from $p_{xy}$. An estimation of  HSIC is given by:
\begin{equation}
\text{HSIC}(Z,\mathcal{F},\mathcal{G}) = \cfrac{1}{(m-1)^2} \textbf{tr} (KH_mLH_m)
\end{equation}
where $K$ and $L$ are matrices containing the evaluation of the reproducing kernel of $\mathcal{F}$ and $\mathcal{G}$ respectively,  and $H_m$ is the centering matrix of size $m$, $H_m = I - \frac{1}{m}\mathbf{1}\mathbf{1}^{\top}$.
\end{definition}
\section{Algorithm}
Let $\mathcal{X}$ be a unit ball in $d$-dimensional space and $X$ contain $n$ observations from $\mathcal{X}$ in form of a $d\times n$ matrix, i.e. $X = [\mathbf{x}_1,\mathbf{x}_2,...,\mathbf{x}_n]$ where each $\mathbf{x}_i \in \mathbb{R}^d$ is a column vector.  According to this definition, $\parallel\mathbf{x}_i\parallel_2\leq 1 \hspace{.3cm} \forall i = \{1,..,n\}$, where $\parallel.\parallel_2$ is the \textit{L}-2 norm of the vector.

Suppose from $n$ samples,  $l$ of them have labels and the rest $u = n- l$ are unlabeled. $X_L$ and $X_U$ contain the set of labeled and unlabeld samples, respectively.  Without loss of generality, assume $X$ is ordered such that the first $l$ samples are labeled, i.e. $X = [ X_L,X_U ]$. Suppose there are also $C$ classes of data points  $\{1,2,...,C\}$. Variable $y_i$ denotes the label of data point $\mathbf{x}_i$ in $X_L$. For data points $\mathbf{x}_j$ in $X_U$, $y_j$ is unknown. The goal of our algorithm is to map the data to a $p$-dimensional space by finding a linear transformation, denoted it by $V$. $V$ is a $d\times p$ matrix where $d$ can be much larger than $p$. Let $\mathbf{z}_i$ be the low-dimensional representation of data point  $\mathbf{x}_i$. Then: $\mathbf{z}_i = V^{\top} \mathbf{x}_i$, where $V^{\top}$ is the transposed of $V$. For the matrix representation form: $Z = V^{\top}X$.

\textbf{Labeled Data:} For the labeled data, we try to find a mapping that maximizes the dependency between low-dimensional data points and the labels, based on the HSIC measure (details in appendix). Therefore, we will have the following objective:
\begin{equation}
\label{eq: obj_lab}
\begin{array}{c}
\arg \max \limits_V \cfrac{1}{(l-1)^2} \textbf{ tr} (Z_L^{\top}Z_L H_l K_l H_l) 
= \arg \max \limits_V \cfrac{1}{(l-1)^2} \textbf{ tr} (X_L^{\top}VV^{\top}X_L H_l K_l H_l)
\end{array}
\end{equation}
where we use linear kernel for the data points in $p$-dimensional space and $K_l$ is a kernel over labels. A kernel commonly used for labels is the delta kernel. Entry $(i,j)$ of a delta kernel is $1$ if $\mathbf{x}_i$ and $\mathbf{x}_j$ have the same label and $0$ otherwise. 
We will use this kernel for labels throughout this paper. 
If we do not impose any constraint on $V$, the function can be unbound. A good choice for the constraint which also guarantees the orthonormality of the basis of the $p$-dimensional space is $V^{\top}V= I$, where $I$ is the identity matrix. By adding this constraint we have:
\begin{equation}
\label{eq: obj_lab_2}
\begin{aligned}
\arg \max \limits_V  \cfrac{1}{(l-1)^2}\textbf{ tr}&(V^{\top}X_L H_l K_l H_lX_L^{\top}V) \hspace{2cm}\text{subject to }  &&V^{\top}V = I
\end{aligned}
\end{equation}
For the sake of simplicity, we do not write the $V^{\top}V= I$ in the next expressions. However, we always consider this constraint in defining objective functions. The objective function in (\ref{eq: obj_lab_2}) can be recast using $X$ and a kernel $K_n$ defined over X. $K_n$ is an  $n \times n$ matrix with all zero entries except the first $l\times l$ block, which is equal to $K_l$. Then, we will have: 
 \begin{equation}
\label{eq: obj_lab_3}
\arg \max \limits_V \cfrac{1}{(n-1)^2}\textbf{ tr}(V^{\top} XH_n K_n H_nX^{\top}V) \\
\end{equation}
\textbf{Unlabeled Data:} The goal here is to find a transformation that preserves the neighborhood between unlabeled data points and their labeled neighbors. We want the unlabeled points to have high similarity with their labeled neighbors in the $p$-dimensional space. This is a rational choice, as a  common assumption in semi-supervised learning is that close points in original space are likely to have same labels. 
If unlabeled data point $\mathbf{x}_i$ and labeled data point $\mathbf{x}_j$ are neighbors in $d$-dimensional space,   then $\mathbf{z}_i$ and $\mathbf{z}_j$ should have high similarity. We measure the similarity between two points $\mathbf{z}_i$ and $\mathbf{z}_j$ by dot product of the centered version of the points dentoed  by  $<\bar{\mathbf{z}}_i,\bar{\mathbf{z}}_j>$. Hence, we can define a function for measuring the similarity of neighboring points: $\max \sum \limits_{ij} w_{ij} <\bar{\mathbf{z}}_i , \bar{\mathbf{z}}_j> = \max \sum \limits_{ij} w_{ij} \bar{\mathbf{z}}_i^{\top} \bar{\mathbf{z}}_j$, where $0 \leq w_{ij}\leq 1$ determines the strength of neighborhood between $\mathbf{x}_i$ and $\mathbf{x}_j$. Note that if both of these points are labeled then $w_{ij}= 0$, as we have already taken care of labeled points in $K_n$. Maximizing this objective function forces points with strong neighborhood (large $w_{ij}$) to have large similarity. The value of $w_{ij}$ between two unlabeled points depend on their similarity in term of their neighborhood. For example, if two unlabeled points have strong neighborhood with labeled points from similar class, then $w_{ij}$ is high. We define an $n\times n$ matrix $W$ that contains $w_{ij}$'s. Based on our definitions here, the first $l\times l$ block of this matrix is all zeros. The objective function  can be written in the following matrix form:
\begin{equation}
\label{eq: unl_mat_form}
\begin{array}{l}
\sum \limits_{ij} w_{ij} \bar{\mathbf{z}}_i^{\top} \bar{\mathbf{z}}_j = \textbf{tr} (\bar{Z}^{\top} \bar{Z} W) = \textbf{tr} (H_nZ^{\top} ZH_n W) =
\textbf{tr} (V^{\top} X H_n W H_n X^{\top} V)
\end{array}
\end{equation}
Therefore, we can also write this objective function similar to (\ref{eq: obj_lab}) by multiplying the trace function to the normalization factor $1/(n-1)^2$ and adding a constraint on $V$.
\begin{equation}
\label{eq: obj_unl}
\arg \max \limits_V  \cfrac{1}{(n-1)^2}\textbf{ tr}(V^{\top}X H_n W H_nX^{\top}V) \\
\end{equation}
Combining (\ref{eq: obj_lab_3}) and (\ref{eq: obj_unl}), we should find mapping $V$ such that the following objective is maximized.
\begin{equation}
\label{eq: obj_ove}
\arg \max \limits_V  \cfrac{1}{(n-1)^2}\textbf{ tr}(V^{\top}X H_n (K_n +W) H_nX^{\top}V)  \\
\end{equation}
The inner matrix, $K_n+W$, is the matrix we needed. 
Elements of $K_n+ W$ show our certainty in similarity of different points in the space. For labeled nodes, we have $0$ and $1$ which indicates absolute certainty. For unlabeled nodes, we have $0 \leq w_{ij} \leq 1$, which is an indicator of our uncertainty. To capture these properties, we define a $C$-dimensional label vector for each data point. For the data point $\mathbf{x}_i$, the label vector is denoted by $\mathbf{y}_i$. If $\mathbf{x}_i$, is labeled then $\mathbf{y}_i$ is an all zero vector except in position $y_i$, which gets value $1$ and it determines the class of $\mathbf{x}_i$. If $\mathbf{x}_i$ is unlabeled, then the $c^{th}$ element of $\mathbf{y}_i$, which we denote it by $y_i^c$, is the probability that $\mathbf{x}_i$ belongs to class $c$, and $\sum_{c=1}^C  y_i^c =1$. To assign this label probabilities, we look at the set of the $k$ nearest labeled neighbors of the unlabeled points $\mathbf{x}_i$. Let us denote this set by $\mathcal{L}_{i,k}$. Then: 
\begin{equation}
\label{eq: label_vec}
y_i^c = \cfrac{f_i^c}{\sum \limits_{c=1}^C f_i^c}\text{ \hspace{.4cm}  where \hspace{.4cm} } f_i^c = \sum \limits_{\substack{\mathbf{x}_j \in \mathcal{L}_{i,k} y_j^c = 1}} \mathcal{S}(\mathbf{x}_i,\mathbf{x}_j)
\end{equation} 
where $\mathcal{S(.,.)}$ is a measure of similarity. As nearby unlabeled points are sharing similar labeled points, they are more likely to have similar label probability vectors as well. 

Now lets look at the dot product of label probability vectors of two points $\mathbf{x}_i$ and $\mathbf{x}_j$, i.e. $\mathbf{y}_i\mathbf{y}_j^{\top}$ ($\mathbf{y}_i$'s are defined as row vectors). If $\mathbf{x}_i$ and $\mathbf{x}_i$ are labeled, this dot product builds elements of Delta kernel matrix, and if one of the points is unlabeled, the dot product builds elements of $W$. Therefore, we can build $K_n+W$ simply by $YY^{\top}$ where $Y$ is an $n\times C$ label matrix. The $i^{th}$ row of $Y$ is $\mathbf{y}_i$, the label vector of $\mathbf{x}_i$. Based on the ordering, we defined for the data points, the first $l$ rows of $Y$ will be corresponding to the labeled points and rest of the rows will be corresponding to the unlabeled data.

Based on the above descriptions, the objective function in (\ref{eq: obj_ove}), is equal to:
\begin{equation}
\label{eq: obj_ove}
\arg \max \limits_V  \cfrac{1}{(n-1)^2}\textbf{ tr}(V^{\top}X H_n YY^{\top} H_nX^{\top}V)  \\
\end{equation}
This is the objective we use to find the $d \times p$ mapping matrix $V$. The columns of the mapping matrix are the eigenvectors corresponding to the top $p$ eigenvalues of $X H_n YY^{\top} H_nX^{\top}$.

At the test time, suppose $X_{ts}$ is a $d\times n_{ts}$ matrix that contains $n_{ts}$ test samples. It is clear that the test points can be mapped to low-dimensional space simply  by: $Z_{ts} = V^{\top}X_{ts}$

\subsection{Kernelized Version}
The advantage of a linear transformation is that it explicitly states the basis of new space as a linear combination of the basis of original space. However, in many applications, a linear transformation is not capable of yielding a good  representation of the data in the new space. Kernel trick is a useful method in these situations, by which, we first implicitly take the data points to a high dimensional RKHS using a non-linear function and then find the low-dimensional representation. An important aspect of our algorithm is its ability to be stated in the kernelized form. 

Based on the representer theorem, the matrix $V$, which we find from (\ref{eq: obj_ove}) can be constructed by a linear combination of functions of data points in the Hilbert space. Let $\phi$ be the function in the Hilbert space. Then $V = \phi(X)\beta$. By plugging this in (\ref{eq: obj_ove}) and replacing $\phi(X)^{\top} \phi(X)$ by the kernel matrix $K_X$, we will have:
\begin{equation}
\label{eq: obj_ker}
\begin{aligned}
&\arg \max \limits_{\beta} &  \cfrac{1}{(n-1)^2}\textbf{ tr}& (\beta^{\top} K_X H_n YY^{\top} H_nK_X \beta)  \\
&\text{subject to }  &&\beta^{\top} K_X \beta = I
\end{aligned}
\end{equation}
ehere $\beta$ is a $n \times p$ transformation matrix. Again, suppose $Q =K_X H_n YY^{\top} H_nK_X$. The solution to (\ref{eq: obj_ker}) that determines $\beta$ is the eigenvectors corresponding to the top $p$ eigenvalues of the generalized eigenvalue problem: $Q\beta = \lambda K_X \beta$. The $p$-dimensional representation of the data is obtained by: $Z = \beta^{\top} K_X$. A popular kernel, which also works very well in our experiments, is the RBF kernel.

For the test data, we should first  compute the kernel similarity between test and training samples. Suppose the entries of the $n\times n_{ts}$ matrix $K_{ts}$ stores the similarities between each pair of training and test data points. Then the $p$- dimensional test data is: $Z_{ts} = \beta^{\top} K_{ts}$.
\section{Bound on the Performance of the SSRL-PL algorithm}
In this section, we derive a bound on the performance of the algorithm. The bound is dependent on the way we assign the probabilities to the unlabeled data points. Let us assume a special case of the SSRL-PL which we call winner take all, or WTA for short. In fact, for any label vector we set the element with the highest probability to one and rest of the elements to zero. Therefore, the $u$ bottom rows of the label matrix $Y$ will also have only 0 and 1. 
Consider the objective in (\ref{eq: obj_ove}). We define the following function:
\begin{equation}
\label{eq: func}
f_X(V,Y) \overset{\bigtriangleup}{=}  \cfrac{1}{(n-1)^2} \textbf{ tr} (V^{\top}X H_n YY^{\top} H_nX^{\top}V) 
\end{equation}

Let ${V^{\dagger}}$ be the solution to (\ref{eq: obj_ove}) when there is $l$ labeled points and $u = n -l$ unlabeled data point in the dataset. Assume $Y_p$ denotes the label matrix in this situation. In addition, consider another situation in which  labels of all data points in $X$ are known. In fact, a completely supervised problem. Let us denote by $Y_n$ the label matrix in this scenario. Suppose $V^*$ is the optimal mapping for the supervised problem., i.e. ${V^{\dagger}} = \arg \max \limits_V f_X(V,Y_p)$ and $V^* = \arg \max  \limits_V f_X(V,Y_n)$.

Our goal is to bound $f_X(V^*,Y_n) - f_X(V^{\dagger},Y_n)$.
In fact, we want to see how much deviation exists between the transformation by $V^*$ and the transformation by ${V^{\dagger}}$. As $f_X(V,Y)$ is a measure of similarity between the labels and the low-dimensional data points, this bound shows the extent to which the low-dimensional representation of the data by ${V^{\dagger}}$ is similar to the real labels of the data points. Note that since $V^*$ is optimal solution for $f_X(V,Y_n)$, this difference is always non-negative.  
\begin{lemma}
\label{lm: middle}
Suppose $X$ is a $d\times n$ matrix of data points  and $Y$ is a $n\times C$ matrix of labels. Based on the definition in (\ref{eq: func})
\begin{equation}
f_X(V,Y) = \cfrac{n^2}{(n-1)^2}\parallel V^{\top}(\overline{XY}-\bar{X}\bar{Y})\parallel_F^2
\end{equation}
where $\parallel.\parallel_F$ is the Frobenius norm of matrix. $\bar{X}_{d\times 1}$ and $\bar{Y}_{1\times C}$ are average of data points and label vectors, respectively, and columns of $\overline{XY}_{d\times C}$ are the weighted average of data points, where weights are columns of $Y$. 
\end{lemma}

Based on the above lemma, we can conclude that: $\arg \max \limits_V f_X(V,Y) = \arg \max \limits_V \sqrt{f_X(V,Y)}$.
$V^{\dagger}$ and $V^*$ are still the maximizers of $ \sqrt{f_X(V,Y_p)}$ and $\sqrt{f_X(V,Y_n)}$, respectively.
As we have bounded $V$ by the constraint $V^{\top}V = I$, the values of $f_X(V,Y)$, and subsequently $\sqrt{f_X(V,Y)}$, are also bounded. Therefore, we can bound the difference of square root of the functions.

We do this to be able to use the properties of the Frobenius norm ($\parallel . \parallel_F$ is a norm, $\parallel .\parallel_F^2$ is not). The following theorem states the bound on difference between square roots. 
\begin{theorem}
\label{thm: final}
Suppose $\mathcal{X}$ is a unit ball in $\mathbb{R}^d$. For $n$ samples drawn iid, according to some probability measure, from $\mathcal{X}$, where the label of only $l$ of them is known and the rest $u$ points are unlabeled, the mapping learned by SSRL-PL algorithm causes at most the following deviation from the mapping that maximizes the HSIC similarity measure between data points and all their revealed real labels. 
\begin{equation*}
\sqrt{f_X(V^*,Y_n)} - \sqrt{f_X({V^{\dagger}},Y_n)} \hspace{.2cm}  \leq \cfrac{2(2+\sqrt{2})u}{n-1} P_e^{\text{WTA}}
\end{equation*}
where $P_e^{\text{WTA}}$ is the error of WTA classifier.
\end{theorem}

As we can see from this theorem, the gap between the two functions vanishes when $u$ is reduced, which shows the consistency of the derived bound. Another important observation about this bound is its independence to dimensionality of original and target space. Therefore, it can be extended to the kernel version as well. Furthermore, suppose that $\hat{V} = \arg \max  \limits_V \lim \limits_{n\ \rightarrow \infty}f_X(V,Y_n)$. In \cite{ashtiani2015}, it has been shown that the deviation of the $f_X(V,Y_n)$ under $\hat{V}$ and $V^*$ is of order $O(1/\sqrt{n})$. This, together with the results of Theorem 3 can yield a generalization bound on SSRL-PL. 
\section{Experiment Results}
In this section, the evaluation of applying the above algorithm on different synthetic and real datasets is presented.  The parameter of the algorithm for each experiment is obtained by leave-one-out cross-validation. We also use RBF kernel similarity in (\ref{eq: label_vec}).

\begin{table*}[!b]
\scriptsize
\label{tbl: comp_UCI}
\caption{Comparison of classification accuracy ($\%$). $\ell/tr$ = portion of the training set that is labeled. The bold numbers show the best results. $p=$ dimensionality of the projected space, $k= $ number of labeled neighbors for each unlabeled data point.}
\begin{center}
\begin{tabular}{lccccccccc}
\hline
Dataset & $l/tr$ & DKSVD & FDDL & LCKSVD2 & OSSDL & S2D2  &  SSRL-PL & $p$ & $k$ 
 \\
\hline
\hline
\noalign{\smallskip}
\multirow{ 2}{*}{\textbf{MNIST-10K}}&	0.1 & 67.18 $\pm$ 1.4	& 74.32$\pm$2.8	& 	69.91$\pm$1.2& 75.15$\pm$1.7 &76.18$\pm$1.5 & \textbf{77.18 $\pm$1.6} & 10 & 5 \\
 									& 0.2 & 70.32$\pm$1.8 & 	79.41$\pm$1.4	& 72.56$\pm$2.2	& 78.52$\pm$1.5& 83.61$\pm$0.9 &  \textbf{85.41$\pm$2.3} & 10 &5\\
\hline
\multirow{ 2}{*}{\textbf{USPS} }	& 0.1 & 60.12$\pm$4.5& 	75.63$\pm$3.6	&75.91$\pm$2.6	& 79.13$\pm$1.3&79.61$\pm$2.4 & \textbf{80.15$\pm$1.9}  & 12 &5\\
 									& 0.2 &66.61$\pm$4.1 & 	80.12$\pm$1.6	&78.64$\pm$1.6	& 81.35$\pm$1.7& \textbf{85.45$\pm$2.1} & 85.31$\pm$2.3  & 12 &5\\
\hline
\multirow{ 2}{*}{\textbf{COIL-20} }	& 0.05 & 52.26$\pm$3.1 & 	68.31$\pm$3.8	& 70.23$\pm$3.1	&81.06$\pm$3.4 & 80.25$\pm$3.8 & \textbf{82.34$\pm$1.2}  & 10 &5\\
 								 	& 0.1&   	56.31$\pm$6.1	&	73.56$\pm$4.1& 76.63$\pm$3.7&86.91$\pm$1.5 & 88.88$\pm$1.0& \textbf{89.71$\pm$0.8}  & 10 &5\\
\hline
\multirow{ 2}{*}{\textbf{Reuters-10K} }& 0.1 & 44.91$\pm$3.6 & 49.81$\pm$3.7		&	55.18$\pm$3.1& 60.21$\pm$1.9& 59.31$\pm$1.8 & \textbf{61.12$\pm$3.1}  & 24 &9 \\
 									&0.2 & 49.32$\pm$1.6& 	57.18$\pm$1.2	&	59.31$\pm$1.7&65.12$\pm$2.3 & 65.18$\pm$3.1 &  \textbf{66.91$\pm$1.2} & 24 &9\\
\hline
\multirow{ 2}{*}{\textbf{UMIST} }	&0.1 & 75.6$\pm$1.3& 80.36$\pm$2.2		&77.33$\pm$2.1	& 79.18$\pm$2.5& 79.65$\pm$1.9 &  \textbf{81.21$\pm$2.3} & 20 &5 \\
 									& 0.2& 79.2$\pm$1.6 & 	83.78$\pm$1.2	&	81.18$\pm$1.3&83.41$\pm$2.1 & 82.11$\pm$2.3 & \textbf{84.31$\pm$2.1}  & 20 &5\\
\hline
\multirow{ 2}{*}{\textbf{SBData} }	& 0.1 & 40.31$\pm$3.9	& 	52.34$\pm$1.2	& 51.23$\pm$2.2	& 49.36$\pm$2.2& 50.87$\pm$2.1 &  \textbf{56.12$\pm$2.6} & 10 &5 \\
 									& 0.2 & 43.69$\pm$3.4 & 	57.36$\pm$2.8	&55.37$\pm$1.6	& 52.34$\pm$2.1&55.62$\pm$1.2 & \textbf{61.74$\pm$1.4 } & 10 &5	\\
\hline
		
\end{tabular}
\end{center}
\label{tbl: comp_ssdl}
\end{table*} 
 
\subsection{Toy Example}
\begin{figure}[!h]
    \centering
    \subfloat[]{{\includegraphics[trim = 14mm 20mm 10mm 20mm,width=2.7cm]{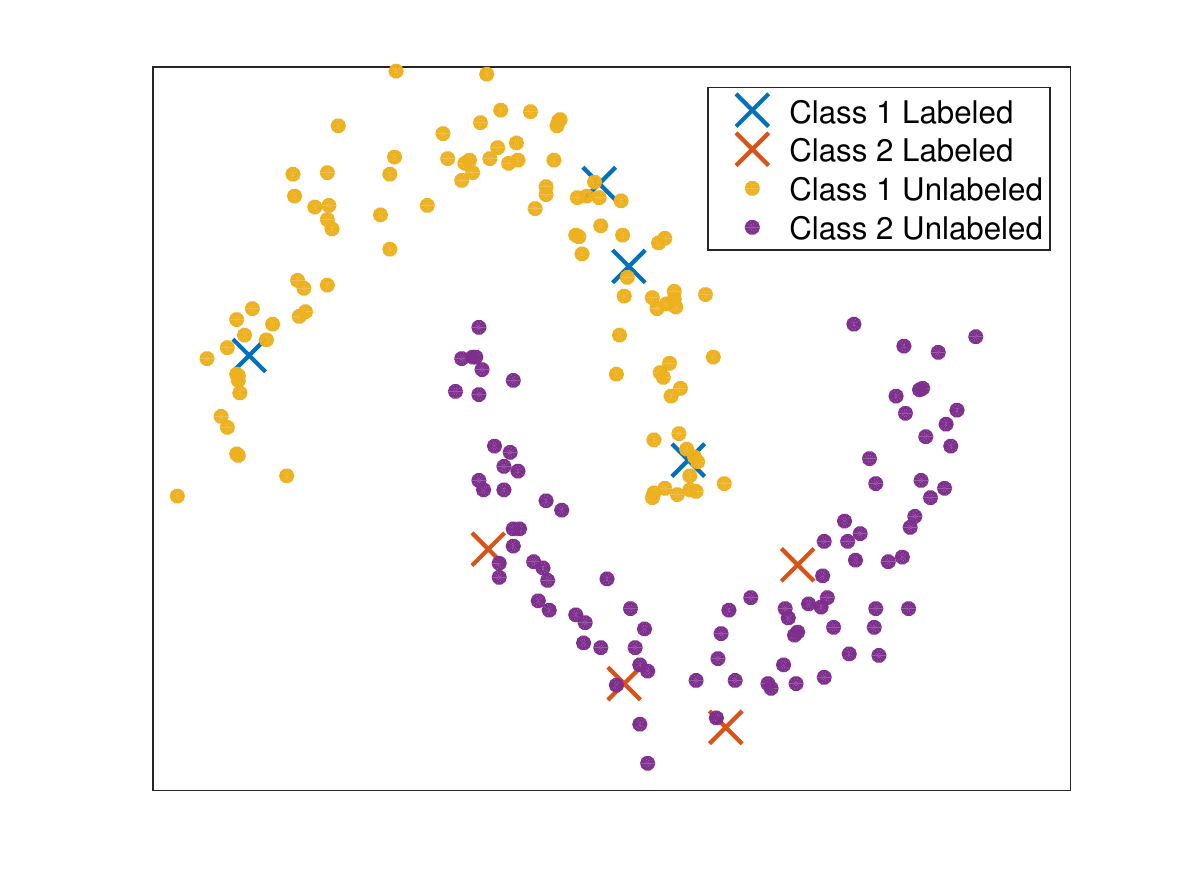} }}
\subfloat[]{{\includegraphics[trim = 14mm 20mm 10mm 20mm,width=2.7cm]{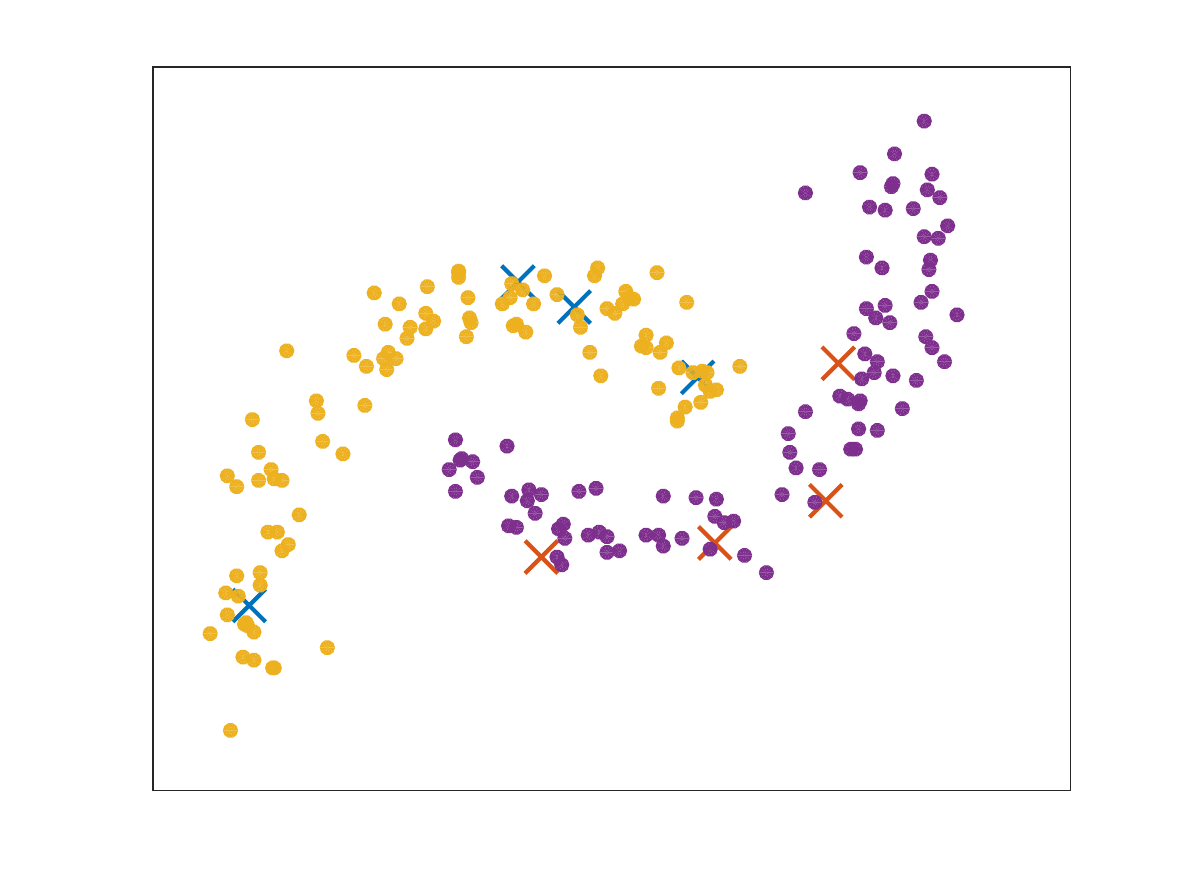} }}
        \subfloat[]{{\includegraphics[trim = 14mm 20mm 10mm 20mm,width=2.7cm]{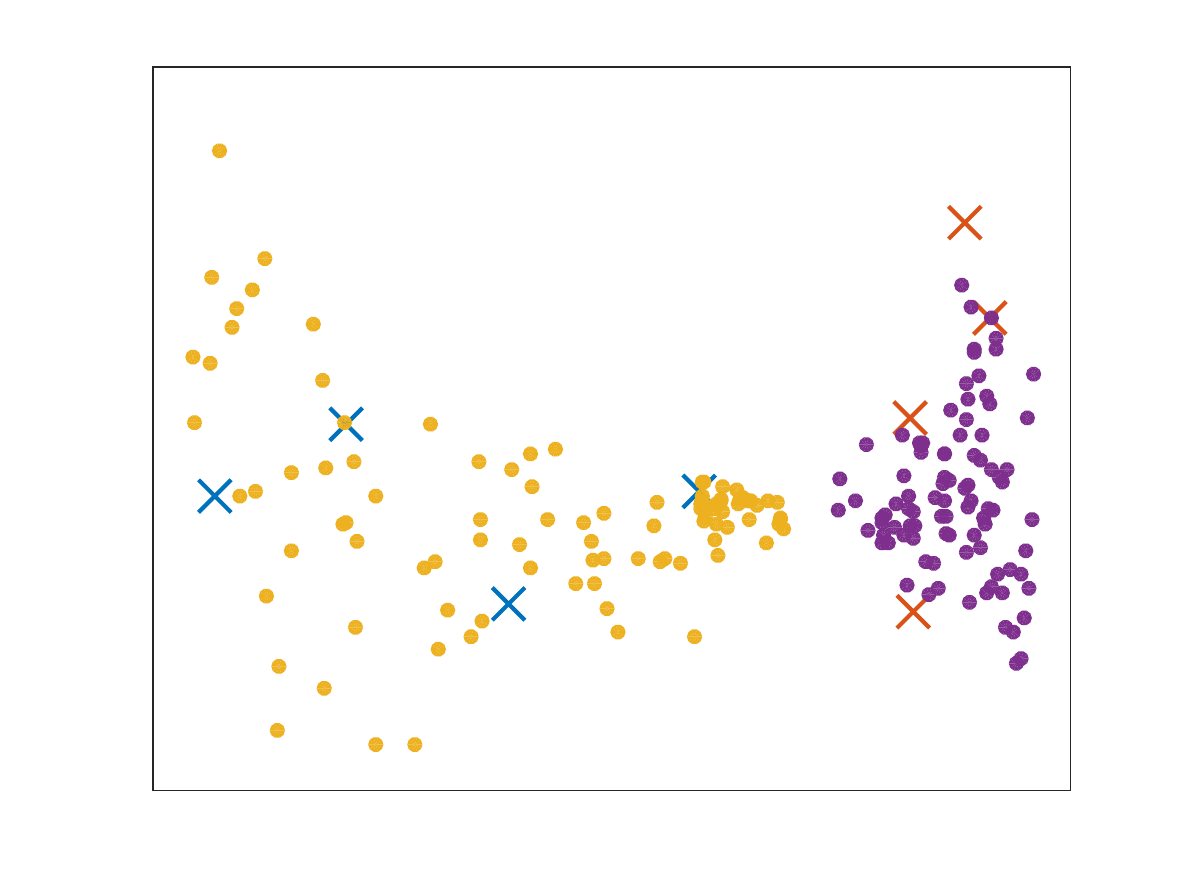} }}
    \caption{(a) Original dataset with 4 labeled data points in each class, (b) SSLR-PL without using kernel $k=1$, (c) SSLR-PL with RBF kernel $\sigma = 0.15$, $k=3$.} 
    \label{fig:twomoons}%
\end{figure}

First, to demonstrate the capabilities of the SSRL-PL algorithm, we apply it on a toy dataset. The two-moon dataset is a well-known for illustrating the effectiveness of an algorithm on a small set of points. The dataset has $200$ samples in two almost balanced classes. Here in Fig. \ref{fig:twomoons}, the results of applying the SSRL-PL algorithm on the dataset is demonstrated, for both kernelized and non-kernelized versions. The number of labeled points in each class is $4$, i.e. $0.04$ of all points.  As it can be easily seen, the algorithm is able to identify the correct labels based the label probability assignments. In the kernelized version, the new representation also provides the ability to classify the points using a linear discriminant.

\subsection{Demonstration and Benchmarks}
Here, we present the results of applying the algorithm on more challenging datasets. The USPS dataset is used to show the generalizabilty of the algorithm and some other datasets from UCI repository are used to show the effectiveness of the algorithm in finding a good representation of data that is suitable for classification, despite the fact the dimensionality of the projected space is much lower than the dimensionality of the original space.

\subsubsection{USPS}
\begin{figure}[!h]
    \centering
    \subfloat[]{{\includegraphics[trim = 10mm 18mm 00mm 30mm,width=4cm]{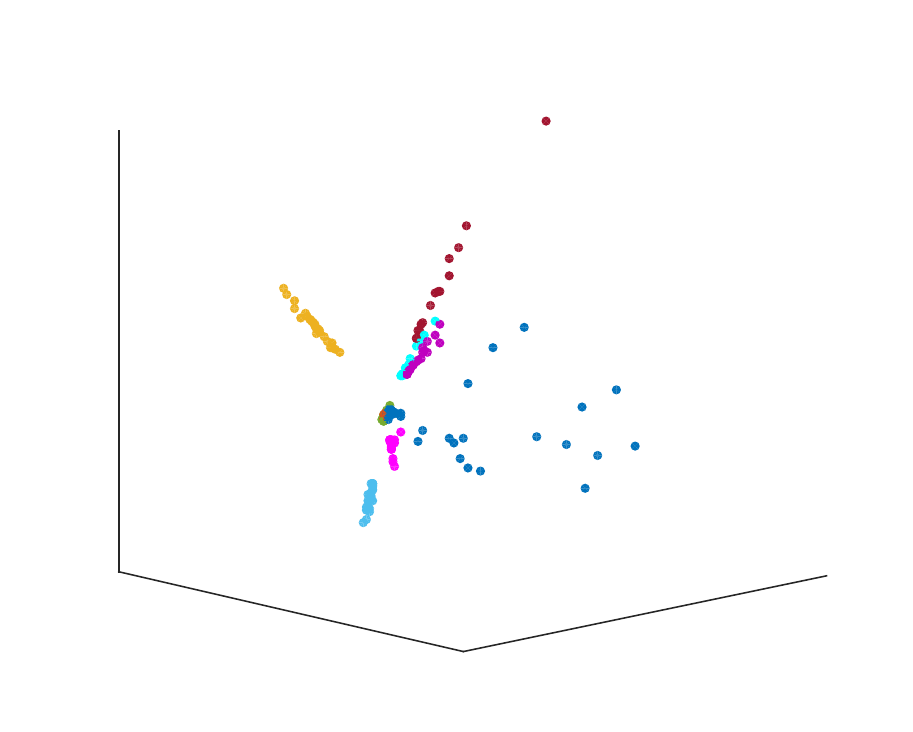} }}
\subfloat[]{{\includegraphics[trim = 10mm 20mm 00mm 30mm,width=4cm]{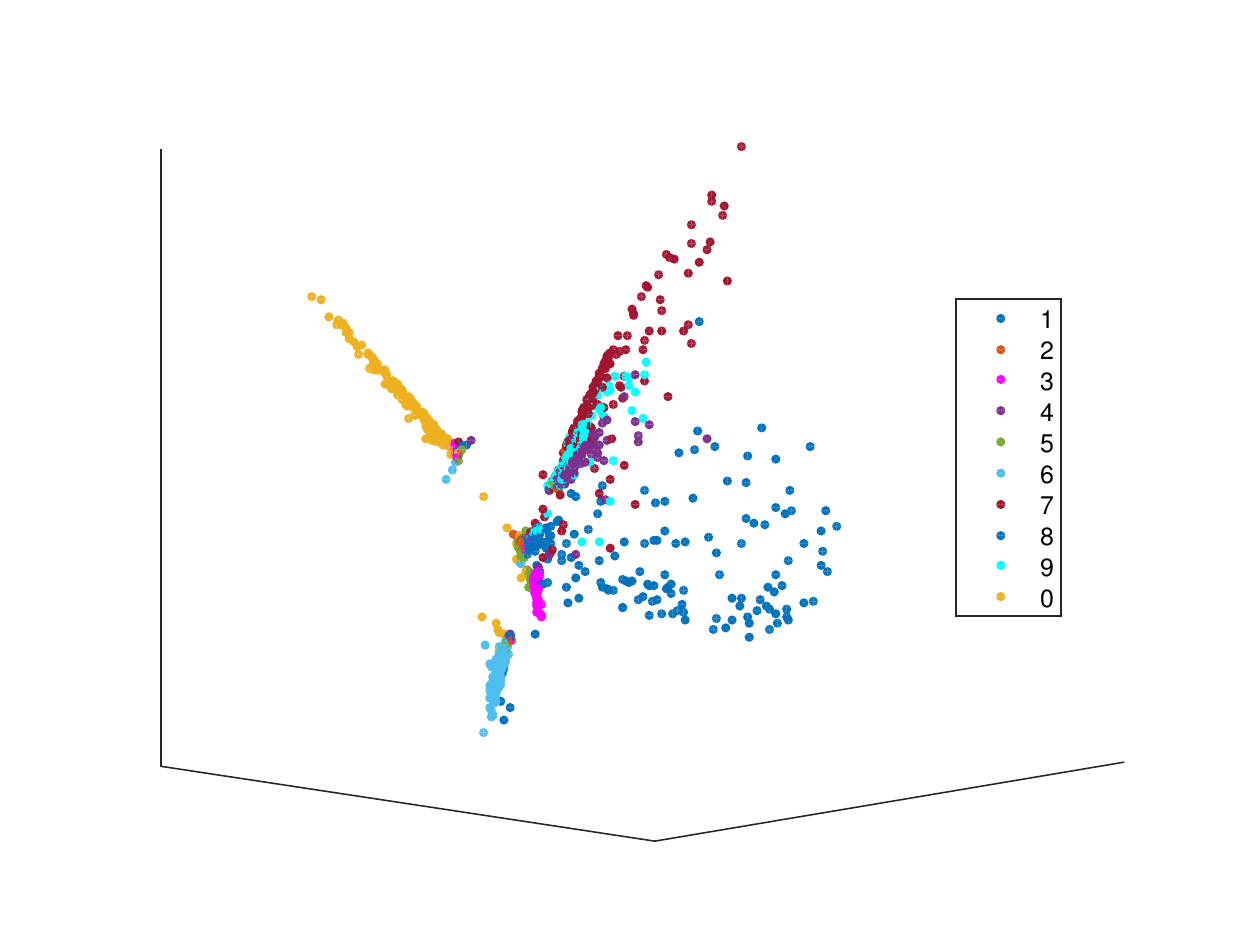} }}%
 \qquad
 
    \caption{(a) Labeled dataset in a 3-dimensional space, (b) Unlabeled and test dataset in a 3-dimensional space, with $k=1$, and $\sigma = 2$.}  
    \label{fig: USPS}%
\end{figure}

USPS hand-written digit dataset consists of $11000$ data points in $10$ classes. The classes are balanced and each of them has $1100$ images of size $16 \times 16$ from hand-written digits $0$ to $9$. Therefore, the dimensionality of samples is $256$. In this experiments, we randomly chose $2000$ samples from them for training and the rest is only used for the testing. The training set is divided into labeled and unlabeled sets. In fact, $10\%$ of the data is labeled. The models is trained by the training set and the obtained transformation matrix, $V$, is applied on both training and test sets.  Figure \ref{fig: USPS} shows the result of applying kernelized SSRL-PL, with RBF kernel, on the dataset. The data is mapped into a three-dimensional space. The left-hand side plot shows the result for only labeled samples of the training set and the right-hand side plot shows the result for both the unlabeled samples of the training set and the test set. We can easily see from this plot that the algorithm is generalizable as its performance on the training set and the large unseen test set is the same.

%
%

\subsubsection{Benchmark datasets}
In \cite{chapelle2006}, multiple benchmarks for the task of semi-supervised learning have been introduced for a fair comparison between algorithms. Datasets can be accessed publicly at \textbf{\texttt{http://olivier.chapelle.cc/ssl-book/benc\\hmarks.html}}. The sets we have used among them are g241c, g241d, and BCI. g241c and g241d both have $1500$ data points and $241$ dimensions, while BCI has $400$ points and $117$ dimensions.  For each dataset, 12 different splits exist, which divide the data into labeled and unlabeled sets. The number of labeled points based on these splits can be either $10$ or $100$. Therefore, the average error rate can be easily reported on these benchmarks. The table below shows the results of applying SSRL-PL on these datasets, according to the provided splits. For comparison, the results of some other algorithms are also reported in the table. These algorithms are LapSVM, LapSVMp\cite{melacci2011}, and Semi-KSC\cite{alzate2012}. The first column of the table, which is titled by $l$, indicates the number of labeled points in the set.
\begin{table}[!h]
\caption{Comparison of the classification error rate of the proposed algorithm with other methods for different datasets. The bold numbers show the best results among these algorithms }

\begin{center}
\small
\begin{tabular}{llccc}
\hline\noalign{\smallskip}
$l$\hspace{-0cm} & Algorithm \ & g241c  & g241d  & BCI
 \\
\noalign{\smallskip}
\hline
\hline
\noalign{\smallskip}
 \multirow{ 4}{*}{\textbf{10} } 			& LapSVM  	& 0.48 $\pm$ 0.02 & 0.42 $\pm$  0.03 & 0.48 $\pm$  0.03  \\
 & LapSVMp  	& 0.49 $\pm$  0.01 & 0.43 $\pm$  0.03 & 0.48 $\pm$ 0.02  \\
			& Semi-KSC	& \textbf{0.42 $\pm$ 0.03} & 0.43 $\pm$  0.04 &0.46 $\pm$ 0.03  \\
 			& \textbf{SSRL-PL} 	&0.43 $\pm$ 0.02&\textbf{0.38 $\pm$ 0.03}& \textbf{0.42 $\pm$ 0.03} \\
\hline
\multirow{ 4}{*}{\textbf{100} } & LapSVM 	& 0.40 $\pm$ 0.06 & 0.31 $\pm$ 0.03 & 0.37 $\pm$ 0.04  \\
			   & LapSVMp 	& 0.36 $\pm$ 0.07 & 0.31 $\pm$ 0.02 & 0.32 $\pm$ 0.02  \\
 				& Semi-KSC	&0.29 $\pm$  0.05 & 0.28 $\pm$ 0.05 &0.22 $\pm$ 0.02 \\
 				& \textbf{SSRL-PL}	&\textbf{0.27 $\pm$ 0.05}&\textbf{0.25 $\pm$ 0.03}&\textbf{0.19 $\pm$ 0.02}
 				 \\
\hline
\end{tabular}
\end{center}
\label{tbl: comp}
\end{table} 
\subsection{Real-world datasets}
Now we examine the performance of the algorithm on six real-world datasets. MNIST-10K is a set of $10000$ images of hand-written digits, which are randomly selected from the MNIST dataset. USPS is also set of images of hand-written digits. UMIST a face recognition dataset. The COIL-20 and SBdata are sets of images of different objects. Reuters dataset \cite{Lewis2004}, contains $810000$ English news stories in different categories. We followed the same procedure in \cite{xie2016} to obtain $10000$ samples from this set in $4$ categories. Other statistics of the datasets are mentioned in table \ref{tbl: datasets}.
\begin{table}[!h]

\caption{Datasets Statistics}
\begin{center}
\small
\begin{tabular}{|l|c|c|c|}
\hline
Datasets Name & \# of points & Dimensionality & \# of classes  \\
\hline
\hline
MNIST-10K 		&  10000	& 784  & 10 \\
\hline
USPS			&  11000	& 256  & 10 \\
\hline
COIL-20			&  1440		& 1024 & 20\\
\hline  
Reuters-10K     &  10000	& 2000 & 4 \\
\hline
UMIST 			&  564		& 750 & 20 \\
\hline
SBData 			&  3192 	& 638 & 40  \\
\hline
\end{tabular}
\end{center}
\label{tbl: datasets}
\end{table}

We compare the performance of the algorithm by multiple dictionary learning algorithms.  Discriminative K-SVD (DKSVD)\cite{zhang2010}, Fisher Discrimination Dictionary Learning (FDDL)\cite{yang2011}, and Label Consistent K-SVD (LCKSVD)\cite{jiang2013} are three supervised dictionary learning algorithms. Also two important semi-supervised dictionary learning algorithm, i.e. OSSDL \cite{zhang2013} and S2D2 \cite{shrivastava2012}. 

We first divide the datasets in two parts, $50\%$ for training and $50\%$ for test. Among the training points we choose $l$ points as labeled and the rest unlabeled such that there is at least one labeled point in each class. We repeat this process $10$ times. Results in table \ref{tbl: comp_ssdl} show the mean and standard deviation of the classification error on the test set. As we can see, the proposed method in this work outperform the other methods. The two other semi-supervised learning algorithms also perform very well. We also include the dimensionality of the target space in the table, which shows that the reduction in dimensionality is significant.

\begin{figure}[!h]
    \centering
    \subfloat[]{{\includegraphics[trim = 10mm 60mm 00mm 60mm,width=4cm]{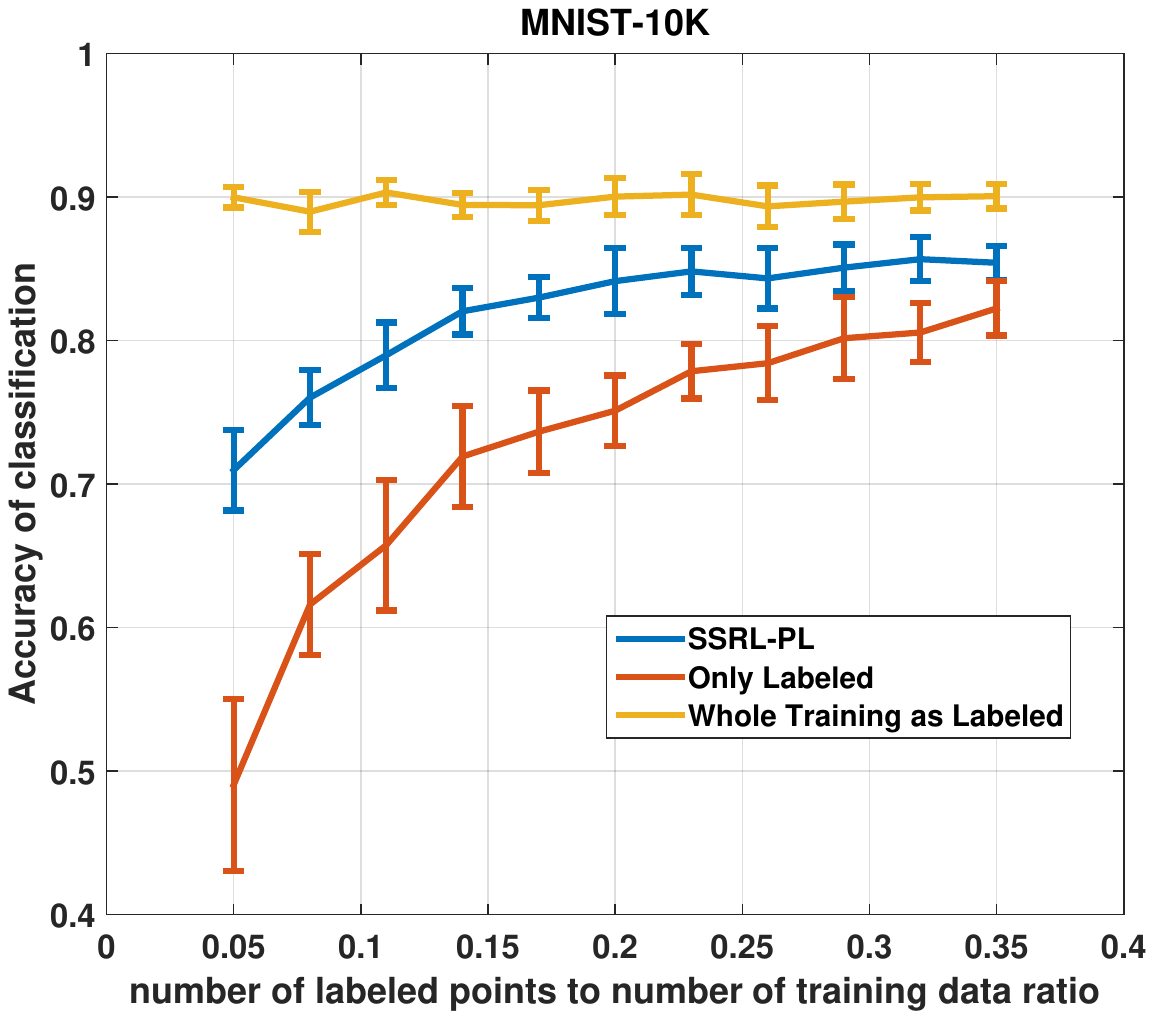} }}
\subfloat[]{{\includegraphics[trim = 10mm 60mm 00mm 60mm,width=4cm]{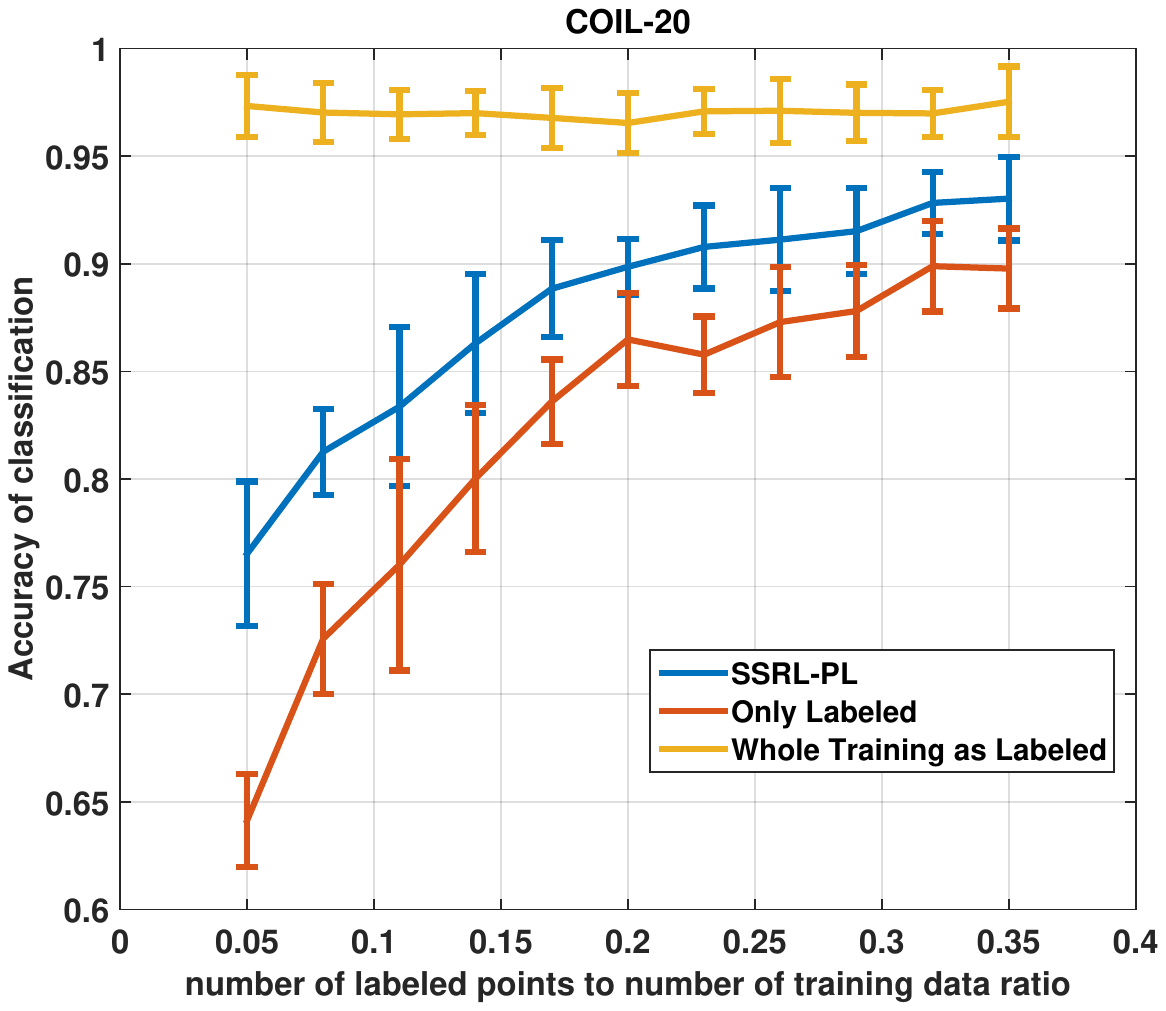} }}%
 \qquad
    \caption{Classification accuracy under three different scenarios (a) MNIST-10K (b) COIL-20}
    \label{fig: compare_whole}%
\end{figure}

For MNIST-10K and COIL-20 we performed another experiment. Again we first divide the datasets in half. Then for different number of labeled points we apply the SSRL-PL algorithm to the resulting training data, for $10$ random splits. We compare the performance of the algorithm with two other scenarios. 1) When only use labeled points to find the mapping $V$, using kernelized version of (\ref{eq: obj_lab_2}). 2) When we use all the labels of the training data and find the mapping $V$, using kernelized version of (\ref{eq: obj_lab_3}). Figure \ref{fig: compare_whole} shows the results of these experiments. As we can see the SSRL-PL performs close to the case when we know all the labels, which shows that the algorithm could convey the label information very well. The fluctuation in the whole labeled results is due to the first random split of dataset to test and train sets.
\section{Conclusion}
We proposed a new algorithm for learning a representation of data when the label information is available for a  small portion of the dataset. The algorithm tries to maximize the similarity between the new representation of data and label set, where the label set for unlabeled data is assigned probabilistically and the similarity measure is HSIC. The effectiveness of the proposed algorithm was evaluated on different datasets. We also derived a bound for the proposed algorithm which can be helpful for seeing if the presence of unlabeled data is constructive or destructive.

In terms of time complexity, the proposed algorithm is equivalent to a standard eigenvalue decomposition problem for symmetric matrices. This problem can be solved efficiently, for example, by singular value decomposition (SVD) methods. However, for faster implementation, using deep autoencoders that are able to estimate eigenvector of their input would be interesting in the future, similar to \cite{banijamali2017fast}. Autoencoders for semi-supervised learning have also been studied in \cite{banijamali2018deep, banijamali2017jade, banijamali2017disentangling}, where two autoencoders are jointly trained to learn a representation for the unlabeled data that contains information about the label.  Alternatively, one can train a network that maximizes the dependency between data points and label vector by optimizing HSIC as its objective function and stochastic gradient descent algorithm. 

\bibliography{semi}
\bibliographystyle{abbrv}

%

\newpage
\appendix

\section{Proofs}
\begin{proof}[Proof of Lemma \ref{lm: middle}]
It is known that: $\sqrt{\textbf{tr}(AA^{\top})}= \parallel A\parallel_F$. We denote the $i^{th}$ column of $Y$ by $\mathbf{y}^i$.
\begin{equation*}
\begin{array}{l}
f_X(V,Y)  = \cfrac{1}{(n-1)^2} \textbf{ tr} (\overbrace{V^{\top}X H_n Y}^A \overbrace{Y^{\top} H_nX^{\top}V}^{A^{\top}}) \\ \\ =\cfrac{1}{(n-1)^2}||\overbrace{V^{\top}XH_nY}^A\parallel_F^2 \\ \\
= \cfrac{1}{(n-1)^2} \parallel V^{\top}\Big (\big [\sum \limits_{i=1}^n (\mathbf{x}_i -\bar{X}) \mathbf{y}_i^1  \hspace{.5cm} \sum \limits_{i=1}^n (\mathbf{x}_i -\bar{X}) \mathbf{y}_i^2 \hspace{.2cm} \\ \\ 
\hspace{3cm} ... \hspace{.2cm} \sum \limits_{i=1}^n (\mathbf{x}_i -\bar{X}) \mathbf{y}_i^C \big]\Big )\parallel_F^2 \\
= \cfrac{1}{(n-1)^2} \parallel nV^{\top}\Big(\hspace{.2cm} \cfrac{1}{n} \big [\sum \limits_{i=1}^n \mathbf{x}_i \mathbf{y}_i^1  \hspace{.4cm} \sum \limits_{i=1}^n \mathbf{x}_i \mathbf{y}_i^2 \hspace{.2cm} ... \hspace{.2cm} \sum \limits_{i=1}^n \mathbf{x}_i \mathbf{y}_i^C \big] \\ \\ \hspace{.7cm} - \cfrac{1}{n}  \big [\sum \limits_{i=1}^n \bar{X}\mathbf{y}_i^1 \hspace{.4cm} \sum \limits_{i=1}^n \bar{X}\mathbf{y}_i^2 \hspace{.2cm} ... \hspace{.2cm} \sum \limits_{i=1}^n \bar{X}\mathbf{y}_i^C\big ]\hspace{.2cm} \Big )\parallel_F^2 \\ \\
= \cfrac{n^2}{(n-1)^2} \parallel V^{\top}\Big( \big [\overline{X \mathbf{y}^1}  \hspace{.4cm} \overline{X \mathbf{y}^2} \hspace{.2cm} ... \hspace{.2cm} \overline{X \mathbf{y}^C} \big] - \bar{X}\bar{Y} \big ]\Big )\parallel_F^2  \\ \\ = \cfrac{n^2}{(n-1)^2}\parallel V^{\top}(\overline{XY}-\bar{X}\bar{Y})\parallel_F^2 
\end{array}
\end{equation*}
\end{proof}

Obtaining the final result for the theorem, needs bounding both $\parallel \bar{Y}_n - \bar{Y}_p\parallel_2$ and $\parallel \overline{XY_n} - \overline{XY_p} \parallel_F$. Lets denote by $\mathbf{e}_i$ the difference between the real label vector and the assigned label vector of point $\mathbf{x}_i$, $\mathbf{e}_i = {\mathbf{y}_n}_i - {\mathbf{y}_p}_i$. For the labeled points $\mathbf{e}_i $ is an all zero vector, for the unlabeled points, if an error happens, the length of $\mathbf{e}_i $ is $\sqrt{2}$. So:
\begin{equation}
\label{eq: bound_1}
\parallel \bar{Y}_n - \bar{Y}_p\parallel_2 = \parallel \bar{\mathbf{e}}\parallel_2 \leq \cfrac{\sqrt{2}u}{n} P_e^{\text{WTA}} 
\end{equation} 

Let $E = Y_n - Y_p$ and $\mathbf{e}^c$ be its $c^{th}$ column. Let also $ne^c$ be the number of errors for class $c$. Note that whether a point in class $c$ misclassified as another class or a point in another class misclassfied as $c$, $ne^c$ increases by one. The bound for $\parallel \overline{XY_n} - \overline{XY_p} \parallel_F$ is then the following:
\begin{equation}
\label{eq: bound_2}
\begin{array}{l}
\parallel \overline{XY_n} - \overline{XY_p} \parallel_F \hspace{.1cm}= \parallel \overline{XE} \parallel_F \hspace{.1cm} \leq \sum \limits_{c=1}^C \parallel \overline{X\mathbf{e}^c}\parallel_2 \\ \\
\hspace{.5cm} = \sum \limits_{c=1}^C \parallel \cfrac{1}{n} \sum \limits_{i=1}^n \mathbf{x}_i ({y_n}_i^c - {y_p}_i^c)\parallel_2 \hspace{.1cm} \\ \\ 
\hspace{.5cm} \leq \cfrac{1}{n} \sum \limits_{c=1}^C ne^c \max \limits_i \parallel \mathbf{x}_i \parallel_2 \hspace{.1cm}\leq \cfrac{1}{n} \sum \limits_{c=1}^C ne^c \leq \cfrac{2u}{n} P_e^{\text{WTA}} 

\end{array}
\end{equation}

\begin{proof}[Proof of Theorem \ref{thm: final}]
Suppose $\epsilon_1 = \overline{XY_n} - \overline{XY_p}$ and $\epsilon_2 = \bar{Y}_n - \bar{Y}_p$. According to the final objective:
\begin{equation*}
\begin{array}{l}
\sqrt{f_X(V^*,Y_n)}\! -\! \sqrt{f_X({V^{\dagger}},Y_n)}= \cfrac{n}{n-1} \times \\ \\
 \Big( \parallel {V^*}^{\top}(\overline{XY_n}-\bar{X}\bar{Y}_n)\parallel_F - \parallel {V^{\dagger}}^{\top}(\overline{XY_n}-\bar{X}\bar{Y}_n)\parallel_F \Big ) \\ \\
= \cfrac{n}{n-1} \Big(\! \parallel {V^*}^{\top}(\overline{XY_p} + \epsilon_1-\bar{X}(\bar{Y}_p + \epsilon_2))\parallel_F \\ \\
\hspace{2cm} - \parallel {V^{\dagger}}^{t}(\overline{XY_p} + \epsilon_1-\bar{X}(\bar{Y}_p + \epsilon_2))\parallel_F  \Big ) \\ \\
\overset{(a)}{\leq}\! \cfrac{n}{n-1} \Big( \parallel {V^*}^{\top}(\overline{XY_p} - \bar{X}\bar{Y}_p)\parallel_F - \parallel {V^{\dagger}}^{t}(\overline{XY_p} - \bar{X}\bar{Y}_p)\parallel_F \\ \\ 
\hspace{2cm}+ \parallel {V^{*}}^{\top} \epsilon_1 \parallel_F +\parallel {V^{\dagger}}^{\top} \epsilon_1 \parallel_F \\ \\ 
\hspace{2cm} + \parallel {V^{*}}^{\top} \bar{X} \epsilon_2\parallel_2 + \parallel {V^{\dagger}}^{\top} \bar{X} \epsilon_2\parallel_2 \Big ) \\ \\
\overset{(b)}{\leq} 
\cfrac{n}{n-1} \Big(
\parallel {V^{*}}^{\top} \epsilon_1 \parallel_F +\parallel {V^{\dagger}}^{\top} \epsilon_1 \parallel_F  \\ \\ 
\hspace{2cm}+ \parallel {V^{*}}^{\top} \bar{X} \epsilon_2\parallel_2 + \parallel {V^{\dagger}}^{\top} \bar{X} \epsilon_2\parallel_2 \Big ) \\ \\
\overset{(c)}{\leq} 
\cfrac{n}{n-1} \Big(
\parallel\epsilon_1 \parallel_F +\parallel \epsilon_1 \parallel_F + \parallel \epsilon_2\parallel_2 + \parallel \epsilon_2\parallel_2 \Big ) \\ \\
\overset{(d)}{\leq} \cfrac{2(2+\sqrt{2})u}{n-1} P_e^{\text{WTA}} \\ \\
\end{array}
\end{equation*}
where inequality $(a)$ comes from triangle inequality, $(b)$ from the fact that ${V^{\dagger}}$ is the maximizer of the $\parallel {V}^{\top}(\overline{XY_p} - \bar{X}\bar{Y}_p)\parallel_F$, $(c)$ from norm properties, and also the fact that orthonormal transformation does not increase the vector length, and finally $(d)$ from (\ref{eq: bound_1}) and (\ref{eq: bound_2}).

\end{proof}

A special case of WTA algorithms is 1-NN. In \cite{drakopoulo1995,nock2001}, a bound on the performance of 1-NN was proposed which can be very helpful for our analysis. Given the underlying class-conditional distribution function is Lipschitz, the probability of error of 1-NN classifier which uses $m$ points as the training is:
\begin{equation}
\label{eq: bound_0}
P_e^{\text{1-NN}} \leq 2P_e^* - \frac{C}{C-1}{P_e^*}^2 + \delta(m)
\end{equation}
where $P_e^*$ is the error of Bayesian classifier, $C$ is the number of classes, and $\delta$ is a penalty factor as a function of number of training points which vanishes as $m \rightarrow \infty$. Using (\ref{eq: bound_0}) we can further bound the algorithm performance which will be independent of the way we assign label to the unlabled data points.

\end{document}